\documentclass[letterpaper, 10 pt, conference]{ieeeconf}

\usepackage{amsmath,amssymb}
\usepackage{graphicx}
\usepackage{booktabs}
\usepackage{multirow}
\usepackage{xcolor}
\usepackage{url}
\usepackage{CJKutf8}
\makeatletter
\let\NAT@parse\undefined
\makeatother
\usepackage[hidelinks]{hyperref}

\makeatletter
\DeclareRobustCommand\selectfont{%
  \ifx\f@linespread\baselinestretch\else
    \set@fontsize\baselinestretch\f@size\f@baselineskip
  \fi
  \ifx\delayed@f@adjustment\@empty\else
    \let\f@shape@saved\f@shape
    \let\f@series@saved\f@series
    \delayed@f@adjustment
    \maybe@load@fontshape
    \ifcsname \f@encoding/\f@family/\f@series/\f@shape \endcsname\else
      \let\f@shape\f@shape@saved
      \let\f@series\f@series@saved
      \let\delayed@merge@font@shape\merge@font@shape
      \let\delayed@merge@font@series\merge@font@series
      \delayed@f@adjustment
      \let\delayed@merge@font@shape\merge@font@shape@without@substitution
      \let\delayed@merge@font@series\merge@font@series@without@substitution
    \fi
    \let\delayed@f@adjustment\@empty
  \fi
  \@forced@seriesfalse
  \xdef\font@name{\csname\curr@fontshape/\f@size\endcsname}%
  \pickup@font
  \font@name
  \expandafter\ifx\csname CJK@\curr@fontshape\endcsname\relax\else
    \CJK@bold@false
    \csname CJK@\curr@fontshape\endcsname
  \fi
  \UseHook{selectfont}%
  \size@update
  \enc@update}
\makeatother

\title{\LARGE \bf
ForeTac: Predictive Contact Guidance for Generative Robot Policies
}

\author{Zhiyuan Chen, Shuai Chen, Haoran Zhou, Yizhe Zhao, Kaiyuan Yang\\
Zhixiang Duan, Shaoqing Xu, Ran Cao, Diyun Xiang, Yangwei You}

\begin{document}
\bstctlcite{ForeTacBSTcontrol}
\begin{CJK*}{UTF8}{gbsn}

\maketitle
\thispagestyle{empty}
\pagestyle{empty}

\begin{abstract}
In contact-rich manipulation, actions that appear equally plausible from vision
can produce different contact outcomes, while reactive control based
on current tactile feedback may respond too late to prevent failure. We present
\emph{ForeTac}, a predictive contact-guidance framework organized as a
\emph{represent--predict--evaluate--guide} loop. ForeTac encodes the current
tactile state, predicts future tactile latent trajectories conditioned on candidate action sequences,
and evaluates each predicted trajectory with a contact-quality objective learned
from successful and failed physical interactions. At inference, gradients
of this objective are backpropagated through the tactile predictor to guide
action generation before execution while keeping all model parameters frozen. We evaluate ForeTac on four real-robot tasks---planar and
curved-surface wiping, card swiping, and fragile-object grasping---with diffusion
and flow-matching policies.
Compared with vision-only Diffusion Policy, ForeTac improves the average task
success rate (SR) from 36.3\% to 71.3\% and the average contact success rate
(CSR) from 20.0\% to 66.3\% across these tasks, and yields gains in both
metrics with a flow-matching policy. Component analyses support the learned
tactile representation, action-conditioned foresight, and generation-time
gradient guidance, while predicted tactile trajectories preserve
contact-quality ordering.
Project page: \url{https://foretac.github.io/}.
\end{abstract}

\section{Introduction}

\begin{figure}[t]
  \centering
  \includegraphics[width=\linewidth]{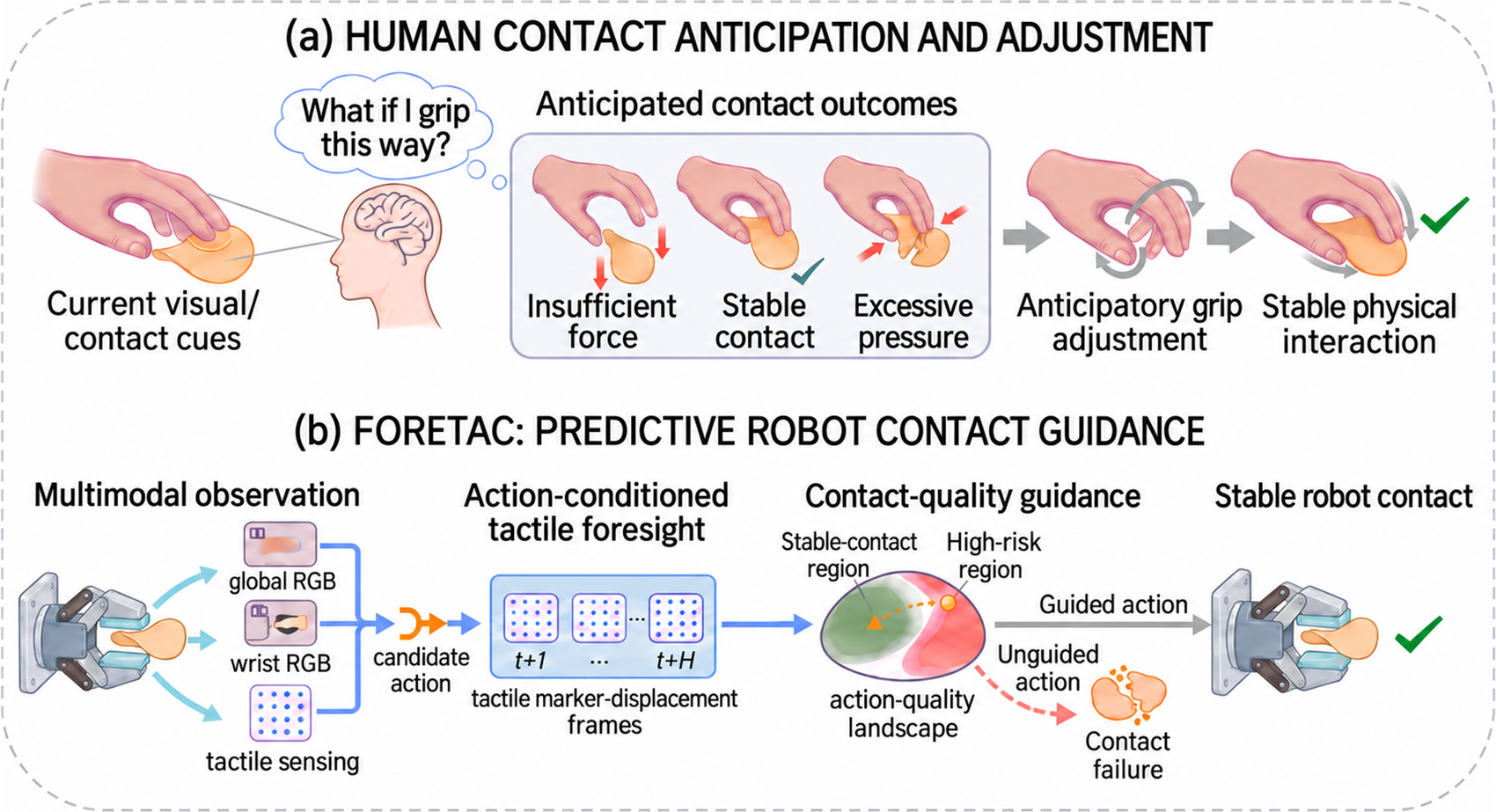}
  \caption{Predictive contact reasoning. (a) Conceptual illustration of human
  contact anticipation: the same current cues can precede different
  action-dependent contact outcomes. (b) ForeTac predicts and evaluates the
  tactile consequences of an evolving candidate action and uses contact-quality
  gradients to guide action generation before execution.}
  \label{fig:teaser}
\end{figure}

Contact-rich manipulation depends not only on where the robot moves but also on
how contact evolves during motion.
Visually similar wiping or swiping actions can maintain contact, lose it, or
apply excessive pressure; delicate grasping actions can secure an object or
irreversibly damage it. Visual observations may obscure local deformation and
slip and do not directly reveal local contact forces. Tactile measurements reveal the
interaction after contact begins, with additional sensing and actuation delay.
When the recovery window is short, a failure may be difficult to prevent by the
time it is observed. Humans adjust grip from predicted fingertip forces before
contact~\cite{flanagan2001sensorimotor}, motivating tactile foresight for robot
policies: predicting and evaluating a candidate action's future contact
consequences while the action is being formed (Fig.~\ref{fig:teaser}).

Learning-based manipulation systems commonly use touch by fusing current
tactile observations into a policy, using tactile prediction as supervision or
context, or applying tactile feedback for reactive correction. Recent studies
use future tactile or physical-interaction prediction as auxiliary supervision
or policy context~\cite{zhang2025tavla,heng2025vitacformer,
higuera2026visuotactilewm,zheng2026omnivta}, while related work uses predicted
contact states for action refinement and contact-aware execution
\cite{tacforesight2026,zhou2026touchworld}. This raises our central question:
how can a generative policy anticipate the contact consequences of an evolving
action, evaluate their quality, and use it to improve the action before execution?

ForeTac augments frozen generative robot policies through a
\emph{represent--predict--evaluate--guide} loop (Fig.~\ref{fig:overview}).
It comprises three components designed for this loop: TacVAE, a spatially
structured tactile representation; Foresight Transformer, an action-conditioned
predictor of multi-step tactile consequences; and a contact-quality scorer that
guides generative action formation. The base policy supplies the task-level
action prior, while ForeTac guides generation toward desirable contact
outcomes.

This paper makes the following contributions:
\begin{itemize}
  \setlength{\topsep}{0pt}
  \setlength{\partopsep}{0pt}
  \item We introduce ForeTac, a predictive contact-guidance framework that
  augments frozen generative robot policies with an outcome-aware tactile
  reasoning loop for contact execution.
  \item We develop a differentiable contact-guidance mechanism that evaluates
  the predicted multi-step tactile consequences of candidate actions and
  propagates contact-quality gradients through the predictor to guide diffusion
  and flow-matching action generation.
  \item Across four real-robot tasks, ForeTac improves SR and CSR over
  vision-only, tactile-fusion, and reactive baselines in the Diffusion Policy
  comparison; it also improves both metrics when attached to a flow-matching
  policy. Ablations and component analyses support the tactile representation,
  action-conditioned foresight, preservation of contact-quality ordering, and
  gradient-based guidance.
\end{itemize}

\section{Related Work}

\subsection{Tactile Representation Learning for Contact-Rich Manipulation}

Large-scale tactile pretraining seeks representations that transfer across
sensors, tasks, and downstream physical-perception problems. Sparsh, AnyTouch,
Sparsh-X, and AnyTouch~2 learn transferable representations from heterogeneous
static or dynamic tactile data~\cite{higuera2024sparsh,feng2025anytouch,
higuera2025sparshx,feng2026anytouch2}. FTP-1 extends this direction toward
shared tactile experts across sensors and embodiments~\cite{yuan2026ftp1}.
Manipulation-oriented methods further preserve spatial or temporal contact
structure: MimicTouch learns from multimodal human tactile demonstrations~\cite{yu2025mimictouch},
while 3D-ViTac unifies visual and tactile observations in a 3-D representation~\cite{huang20243dvitac}.

These studies motivate tactile representations that preserve local contact
structure and temporal evolution while remaining tractable for downstream
control. ForeTac uses a compact spatial latent that preserves marker-field structure and
provides a shared interface for tactile prediction and contact-quality
evaluation.

\subsection{Predictive Tactile Modeling for Robot Manipulation}

Predictive tactile models estimate how an action changes physical interaction
before the resulting contact is observed. Early work learned latent tactile
dynamics for servoing~\cite{sutanto2019latent}, predicted tactile sequences for
tactile-goal model-predictive control~\cite{tian2019manipulation}, or modeled
action-conditioned tactile changes such as slip~\cite{mandil2022action}.
Subsequent visuo-tactile models extended prediction to object-interaction
planning~\cite{zhang2023visualtactile,ai2024robopack} and to joint future
visual-tactile states, action trajectories, or contact subgoals~\cite{
higuera2026visuotactilewm,zheng2026omnivta,zhang2026contactworld,
yuan2026vtam,zhou2026touchworld,lou2026dreamtac}.

Recent methods differ in their prediction targets and conditioning interfaces:
future force or tactile signals can provide auxiliary supervision
\cite{zhang2025tavla,li2025adaptac}, tactile tokens or latent trajectories can
condition a policy or provide an anticipatory prior
\cite{heng2025vitacformer,tacforesight2026}, and draft-action-conditioned
future tactile prediction can support action refinement~\cite{ye2025dreamtacvla}.
ForeTac predicts the multi-step tactile consequences of each evolving candidate
action as a temporally aligned latent trajectory for contact-quality evaluation;
the resulting evaluation guides generative action updates.

\subsection{Inference-Time Guidance for Generative Robot Policies}

Diffusion and flow-matching policies produce actions by iteratively evolving a
generative state~\cite{ho2020ddpm,chi2023diffusionpolicy,lipman2023flowmatching,
black2025pi05}. Classifier guidance modifies this process using the gradient of
an external objective~\cite{dhariwal2021diffusion}. Robot-policy extensions
have used human interaction constraints~\cite{wang2025itps}, predictive
world-model objectives~\cite{qi2026gpc}, and active dynamic guidance
\cite{du2025dynaguide}. These methods either rank sampled candidates or update
the generative state with an external gradient. DynaGuide differentiates a
goal-conditioned cost over predicted visual outcomes~\cite{du2025dynaguide}.
TouchGuide evaluates the compatibility of current visual-tactile observations
and an action estimate during diffusion or flow-matching inference
\cite{touchguide2026}. Hierarchical visuo-tactile steering combines visual
mode selection with short-horizon diffusion editing guided by text-conditioned
rewards on predicted tactile latents~\cite{vitalsteering2026}.

ForeTac evaluates each candidate action's predicted multi-step tactile
trajectory with a contact-quality objective learned from successful and failed
interactions, then propagates its gradient through the predictor to guide
diffusion and flow-matching action generation.

\section{Method}

\begin{figure*}[t]
  \centering
  \includegraphics[width=\textwidth]{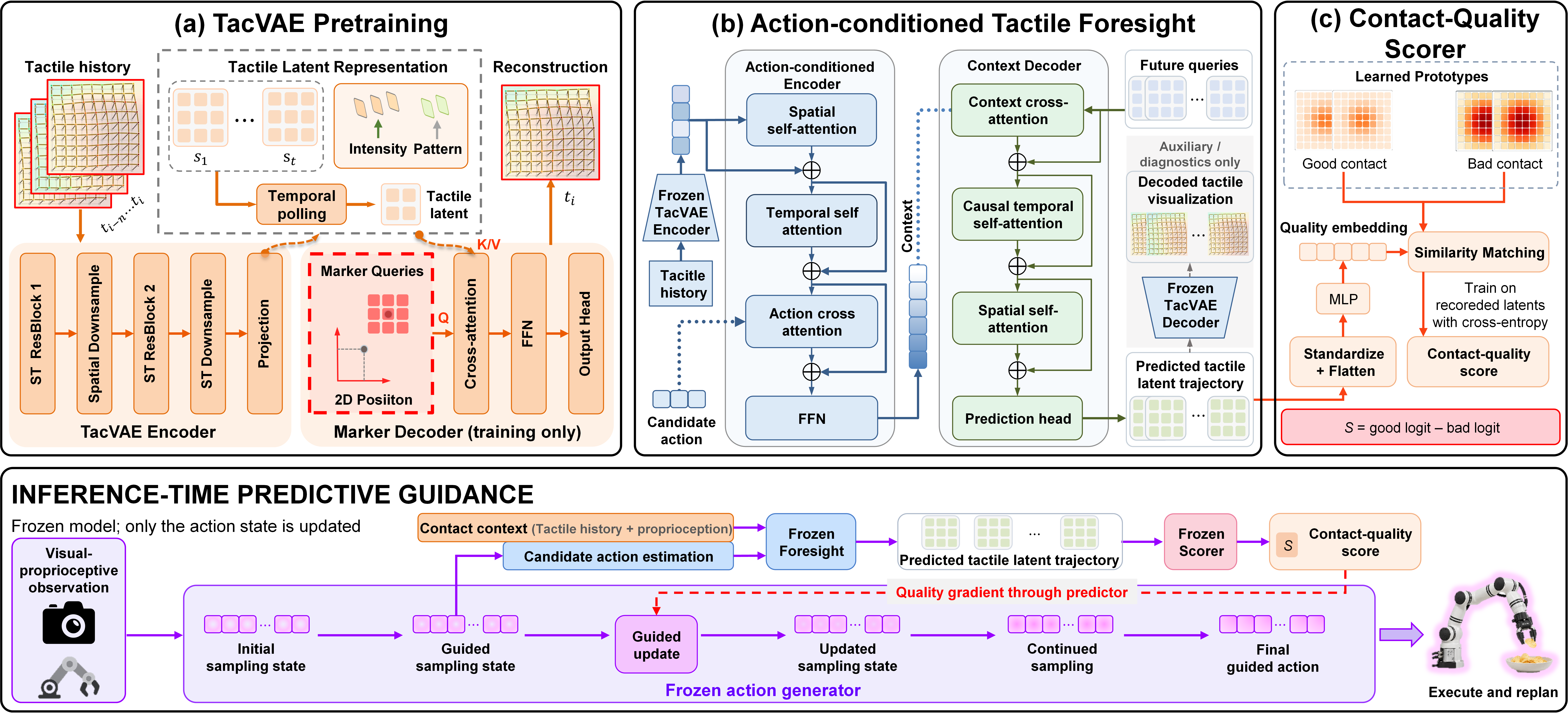}
  \caption{ForeTac overview. The upper panels show tactile representation
  learning, action-conditioned future tactile prediction, and contact-quality
  scoring of predicted tactile latents. The lower panel shows inference-time
  guidance, where predicted contact quality updates the guided sampling state.}
  \label{fig:overview}
\end{figure*}

\subsection{Overview and Formulation}

At time $t$, the frozen base policy receives the visual--proprioceptive
observation
\begin{equation}
  \mathbf{o}_t^\pi = \{I_t^g, I_t^w, \mathbf{q}_t\},
\end{equation}
where $I_t^g$ and $I_t^w$ denote global and wrist RGB images, and
$\mathbf{q}_t$ is the current robot configuration. ForeTac separately forms
the contact context
\begin{equation}
  \mathbf{c}_t = \{\mathbf{q}_t,\mathbf{m}_{t-W+1:t}\},
\end{equation}
where $\mathbf{m}_{t-W+1:t}$ is a history of tactile marker displacements, with
$W=8$ in our implementation. The frozen base policy with parameters $\theta$
maintains an intermediate action state $\mathbf{x}_t^k$ at sampling step $k$ and uses
$h_\theta$ to map it to a provisional clean $H_a$-step action chunk,
\begin{equation}
  \hat{\mathbf{a}}_{t:t+H_a-1}^{0}
  = h_\theta(\mathbf{x}_t^k, k, \mathbf{o}_t^\pi).
  \label{eq:clean_action_estimate}
\end{equation}
The superscript $0$ denotes this clean-action estimate. In the diffusion
derivation below, we suppress the fixed observation time $t$, write
$\mathbf{x}^k$ for $\mathbf{x}_t^k$, and denote the corresponding candidate
action chunk by $\hat{\mathbf{x}}_k^0$; $\hat{\mathbf{x}}_s^0$ is the analogous
notation for flow matching.

Given the contact context and this prospective action sequence, the
action-conditioned tactile predictor forecasts a temporally aligned trajectory
of TacVAE latent states:
\begin{equation}
  \hat{\mathbf{z}}_{1:H}
  = F_\phi(\mathbf{c}_t, \hat{\mathbf{a}}_{t:t+H_a-1}^{0}).
  \label{eq:foresight_contract}
\end{equation}
Here, $F_\phi$ is the action-conditioned Foresight predictor with parameters
$\phi$. $H$ denotes the tactile prediction horizon. We use $H=H_a=16$ by default,
with each predicted tactile frame aligned to the corresponding future action
step. The relative index $h=1,\ldots,H$ identifies future frames relative to the
current observation. The recorded latent targets $\mathbf{z}_{1:H}$ are obtained
by encoding the corresponding marker windows with the frozen TacVAE. The
scorer $S_\psi$ with parameters $\psi$ evaluates
the predicted latent trajectory, and its gradient is propagated through $F_\phi$ to
the intermediate generative state.

ForeTac combines tactile representation, action-conditioned tactile foresight,
and predictive contact guidance in the
\emph{represent--predict--evaluate--guide} loop shown in
Fig.~\ref{fig:overview}. At deployment, the base generative policy, TacVAE,
Foresight, and contact-quality model remain frozen; only the generative
sampling state is updated. The base policy supplies the task-level action
prior, while the scorer provides the quality objective for action formation.
We instantiate this interface with
Diffusion Policy~\cite{chi2023diffusionpolicy} and the flow-matching
$\pi_{0.5}$ policy~\cite{black2025pi05}; their backbone-specific updates are
given in Sec.~\ref{sec:predictive_contact_guidance}, with Diffusion Policy
following the classifier-guidance principle~\cite{dhariwal2021diffusion}.

\subsection{Tactile Representation}

As shown in Fig.~\ref{fig:overview}(a), TacVAE encodes recent contact while
preserving its spatial layout using an eight-frame history
\(\mathbf{m}_{t-W+1:t}\), where each frame is a
$9\times9\times2$ marker-displacement grid. Causal 3-D convolutions and
spatio-temporal residual blocks aggregate only present and past observations,
followed by spatial and spatio-temporal downsampling. The resulting latent
volume has four encoded temporal states. A learned temporal query aggregates
these states into the current tactile latent map
\(\mathbf{z}_t\in\mathbb{R}^{16\times3\times3}\).

The latent projection explicitly decomposes tactile information into an
intensity component and complementary contact-pattern features.

The frozen marker-reconstruction decoder $D_{\mathrm{tac}}$ maps the nine spatial latent tokens back to the
$9\times9\times2$ displacement field using position-aware queries and
cross-attention. It provides reconstruction and auxiliary marker-space
supervision during training but is bypassed at deployment. The
downstream foresight and contact-quality pipeline uses the complete
$16\times3\times3$ latent map (144 dimensions), and the deployment scorer
receives the future latent trajectory produced by Foresight. The TacVAE
training objective is
\begin{equation}
\mathcal{L}_{\mathrm{TacVAE}}=
\mathcal{L}_{\mathrm{rec}}+\lambda_{\mathrm{dir}}\mathcal{L}_{\mathrm{dir}}
+\lambda_{\mathrm{KL}}\mathcal{L}_{\mathrm{KL}},
\label{eq:tacvae_loss}
\end{equation}
where the terms denote marker reconstruction, displacement-direction
consistency, and variational regularization~\cite{kingma2014autoencoding}, respectively. We use
$\lambda_{\mathrm{dir}}=0.2$ and $\lambda_{\mathrm{KL}}=10^{-6}$.

\subsection{Action-Conditioned Tactile Foresight}

Given the current contact context and a candidate action sequence, the
Foresight Transformer predicts the corresponding multi-step tactile trajectory
according to Eq.~\eqref{eq:foresight_contract} and
Fig.~\ref{fig:overview}(b). Its input interface
contains the current TacVAE-encoded tactile tokens, proprioceptive state, and
candidate action chunk. Visual observations remain in the base-policy branch.

The frozen TacVAE converts the current tactile history into nine spatial latent
tokens. These tokens, the proprioceptive state, and the candidate action chunk
are embedded in a 512-dimensional space and processed by a three-layer
action-conditioned Transformer~\cite{vaswani2017attention}. Each context layer
combines within-frame spatial attention, temporal attention across the history
(when at least two frames are available), and cross-attention from observation
tokens to the candidate-action chunk. Learned queries indexed by future time
and tactile cell retrieve this aligned context and generate the multi-step
latent trajectory in one forward pass, with causal information flow across
predicted times and spatial interaction within each predicted map. The main
configuration predicts $H=16$ future steps, each represented by a
$16\times3\times3$ (144-dimensional) tactile latent map.

During Foresight training, the frozen TacVAE supplies the latent targets. We train
the foresight model with
\begin{align}
  \mathcal{L}_{\mathrm{fs}}
  &= \mathcal{L}_{\mathrm{latent}}
  + \lambda_m\mathcal{L}_{\mathrm{marker}}
  + \lambda_{\Delta}\mathcal{L}_{\Delta}, \\
  \mathcal{L}_{\mathrm{latent}}
  &= \operatorname{SmoothL1}(\hat{\mathbf{z}}_{1:H},\mathbf{z}_{1:H})
  + \operatorname{SmoothL1}(\hat{\mathbf{z}}_H,\mathbf{z}_H), \\
  \mathcal{L}_{\mathrm{marker}}
  &= \operatorname{SmoothL1}\!\left(
  D_{\mathrm{tac}}(\hat{\mathbf{z}}_{1:H}),\mathbf{m}_{1:H}\right), \\
  \mathcal{L}_{\Delta}
  &= \operatorname{SmoothL1}\!\left(
  \Delta\hat{\mathbf{z}}_{1:H-1},\Delta\mathbf{z}_{1:H-1}\right),
  \label{eq:foresight_loss}
\end{align}
where $D_{\mathrm{tac}}$ denotes the frozen decoder described above,
$\mathbf{m}_{1:H}$ denotes the recorded marker targets aligned with these future
steps, and $\Delta\hat{\mathbf{z}}_h=\hat{\mathbf{z}}_{h+1}-\hat{\mathbf{z}}_h$
and $\Delta\mathbf{z}_h=\mathbf{z}_{h+1}-\mathbf{z}_h$. We set
$\lambda_m=0.3$ and
$\lambda_{\Delta}=0.5$; the additional final-step latent term has unit weight.

\subsection{Predictive Contact Guidance}
\label{sec:predictive_contact_guidance}

\subsubsection{Contact Quality Objective}

For each task, we independently train a contact-quality model from recorded
physical interactions labeled as good or bad~\cite{tian2026badbehavior}, as illustrated in
Fig.~\ref{fig:overview}(c). During training, the frozen
TacVAE encodes recorded future marker trajectories into tactile latent chunks.

The predicted latent chunk is standardized with fixed per-dimension mean and
standard-deviation statistics from scorer training, then flattened and passed
through a three-layer MLP to obtain a unit-norm 128-D quality embedding
$\mathbf{e}$. Let $\mathbf{p}_c$
denote one of two learned, unit-norm prototypes, with
$c\in\{\mathrm{good},\mathrm{bad}\}$. The corresponding quality logits are
\begin{equation}
  \ell_c(\hat{\mathbf{z}}_{1:H})=
  \frac{\mathbf{e}^{\mathsf T}\mathbf{p}_c}{\tau}+b_c,
  \label{eq:quality_logits}
\end{equation}
where $\tau$ is a temperature and $b_c$ is a learned outcome bias. We train the
scorer with cross-entropy over the two outcome labels, good and bad. For guidance,
we define the contact-quality objective as the good--bad logit margin
\begin{equation}
  S_\psi(\hat{\mathbf{z}}_{1:H})
  =\ell_{\mathrm{good}}(\hat{\mathbf{z}}_{1:H})
  -\ell_{\mathrm{bad}}(\hat{\mathbf{z}}_{1:H}).
  \label{eq:quality_margin}
\end{equation}
The learned margin encodes a preference for good contact over bad contact,
and its gradient through the predictor supplies the guidance signal during
generation.

At deployment, each candidate action chunk is evaluated through its predicted
tactile latent trajectory:
\begin{equation}
  \mathbf{a}\longrightarrow
  F_\phi(\mathbf{c}_t,\mathbf{a})\longrightarrow S_\psi.
  \label{eq:quality_gradient_path}
\end{equation}
Here, $\mathbf{a}$ denotes the clean candidate action chunk, instantiated as
$\hat{\mathbf{x}}^{0}_k$ for diffusion sampling and $\hat{\mathbf{x}}^{0}_s$ for
flow matching. This path returns the contact-quality gradient to the generative
sampler.

\subsubsection{Guidance during Generative Sampling}

\paragraph{Diffusion Policy}
Let $\mathbf{x}^k$ be the normalized action state at diffusion step $k$ and
$\epsilon_\theta$ the frozen noise predictor. During quality backpropagation,
the noise-prediction output is held fixed; gradients act on the current
generative state through the clean-action estimate. The corresponding clean-action
estimate is
\begin{equation}
  \hat{\mathbf{x}}^{0}_k=
  \frac{\mathbf{x}^k-\sqrt{1-\bar{\alpha}_k}\,
  \operatorname{sg}\!\left[\epsilon_\theta(\mathbf{x}^k,k,\mathbf{o}_t^\pi)\right]}
  {\sqrt{\bar{\alpha}_k}}.
  \label{eq:diffusion_clean_estimate}
\end{equation}
Here, $\bar{\alpha}_k$ denotes the cumulative signal-retention coefficient at
diffusion step $k$, and $\operatorname{sg}[\cdot]$ treats the noise-prediction
output as constant when computing the quality gradient.
Using this estimate as the candidate action for the foresight model, we compute
\begin{equation}
  Q_k=S_\psi\!\left(F_\phi(\mathbf{c}_t,\hat{\mathbf{x}}^{0}_k)\right),
  \qquad
  \mathbf{g}_k=\nabla_{\mathbf{x}^k}Q_k.
  \label{eq:diffusion_quality_gradient}
\end{equation}
The normalized quality gradient then updates the state:
\begin{equation}
  \widetilde{\mathbf{x}}^{k}=
  \mathbf{x}^k+\eta_k\frac{\mathbf{g}_k}
  {\lVert\mathbf{g}_k\rVert_2+\epsilon}.
  \label{eq:diffusion_guidance}
\end{equation}
Here $\eta_k$ is the guidance step size and $\epsilon$ is a numerical
stabilizer.
The updated state is then passed to the ordinary diffusion scheduler. Guidance
is applied only at the configured subset of sampling steps, and the executed
action chunk is subsequently replanned by the receding-horizon controller.

\paragraph{Flow-Matching Policy}
For a flow-matching policy such as $\pi_{0.5}$, let $\mathbf{x}_s$ denote the current
flow state and $v_\theta(\mathbf{x}_s,s,\mathbf{o}_t^\pi)$ the frozen velocity field. As in
the diffusion case, the velocity is treated as constant while computing the
quality gradient. With
the flow integrated from $s=1$ toward $s=0$, the provisional action estimate is
\begin{equation}
  \hat{\mathbf{x}}^{0}_s=\mathbf{x}_s-s\,\operatorname{sg}\!\left[
  v_\theta(\mathbf{x}_s,s,\mathbf{o}_t^\pi)\right].
  \label{eq:flow_clean_estimate}
\end{equation}
We evaluate the same predicted-contact objective and obtain its gradient with
respect to the current flow state:
\begin{equation}
  Q_s=S_\psi\!\left(F_\phi(\mathbf{c}_t,\hat{\mathbf{x}}^{0}_s)\right),
  \qquad
  \mathbf{g}_s=\nabla_{\mathbf{x}_s}Q_s.
  \label{eq:flow_quality_gradient}
\end{equation}
With $\gamma_s$ denoting the flow-guidance step size, a normalized proposal is
formed as
\begin{equation}
  \widetilde{\mathbf{x}}_s=\mathbf{x}_s+\gamma_s
  \frac{\mathbf{g}_s}{\lVert\mathbf{g}_s\rVert_2+\epsilon}.
  \label{eq:flow_guided_proposal}
\end{equation}
The velocity is recomputed at the updated state, and the flow proceeds with its
ordinary Euler step:
\begin{equation}
  \mathbf{x}_{s+\Delta s}=\widetilde{\mathbf{x}}_s+\Delta s\,
  v_\theta(\widetilde{\mathbf{x}}_s,s,\mathbf{o}_t^\pi),
  \qquad \Delta s=-1/N.
  \label{eq:flow_guidance}
\end{equation}
Here $s$ is the flow time and $N$ is the number of ordinary flow steps. Thus the
quality gradient modifies the
current flow state rather than being added directly to the velocity field.

\section{Experiments}
\label{sec:experiments}

\begin{figure*}[t]
  \centering
  \includegraphics[width=\textwidth]{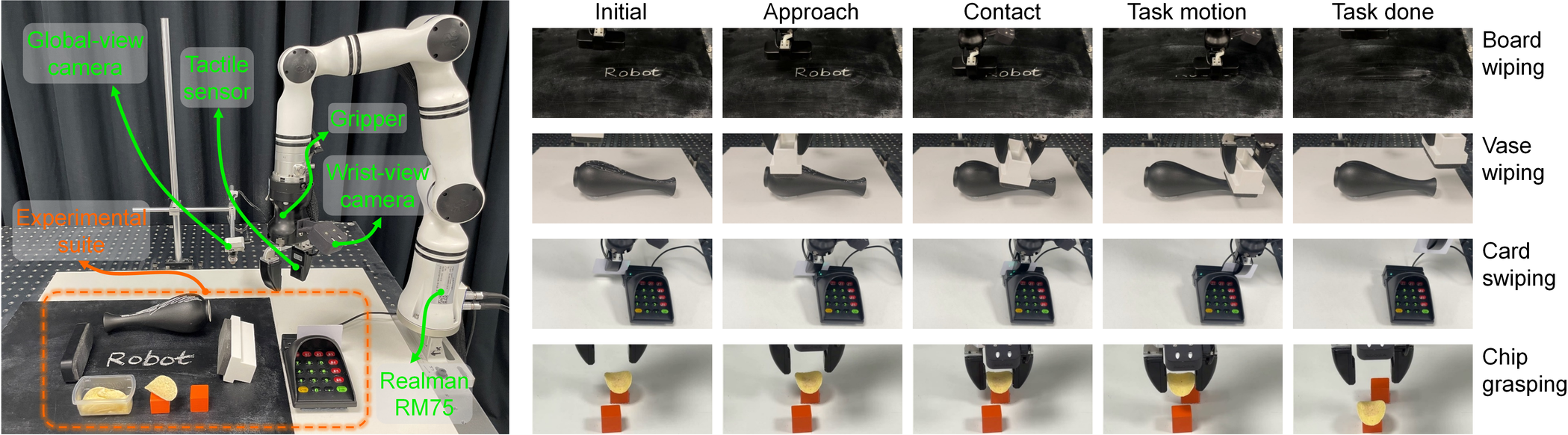}
  \caption{Experimental platform and real-robot task execution. Left: the
  experimental setup. Right: execution snapshots for the four tasks, from the
  initial state to task completion.}
  \label{fig:tasks}
\end{figure*}

\subsection{Experimental Setup}

\subsubsection{Robot Platform}

Experiments use a Realman RM75 arm with a Robotiq Hand-E parallel gripper, a
Vitai GF225 optical visuotactile sensor, and two Intel RealSense D405 RGB
cameras (Fig.~\ref{fig:tasks}); all compared methods use the same platform and
tactile instrumentation.

\subsubsection{Tasks}

Figure~\ref{fig:tasks} summarizes four real-robot tasks: planar board wiping
for contact establishment and maintenance, curved-surface vase wiping for
contact transfer, constrained card swiping with continuous contact and
alignment, and fragile chip grasping with a narrow force range between slip or
drop and damage.

Task definitions and evaluation criteria are fixed before evaluation;
the common trial and reset protocol is specified below and shared by all
compared methods.

\subsubsection{Baselines}

The primary comparison uses the shared task protocol and includes vision-only
\textbf{DP}~\cite{chi2023diffusionpolicy} and \textbf{RDT}~\cite{liu2024rdt},
current-touch \textbf{DP + Tactile Concat}, reactive \textbf{RDP}~\cite{xue2025reactive},
and predictive-touch \textbf{DP + ForeTac}. DP is the vision-only backbone;
DP + Tactile Concat tests direct current-tactile fusion; RDP provides reactive
tactile feedback; and DP + ForeTac predicts candidate-action consequences and
uses their contact quality during generation. The flow-matching $\pi_{0.5}$
policy~\cite{black2025pi05} is evaluated with and without ForeTac in the second
block of Table~\ref{tab:main_results}; auxiliary tactile prediction and
predictive reranking are treated as ForeTac ablations.

\subsubsection{Evaluation Protocol and Metrics}

We report task success rate (SR), the fraction achieving the task-specific
objective, and contact success rate (CSR), the fraction also satisfying
predefined contact criteria:
\begin{equation}
  \mathrm{CSR}=\frac{1}{N}\sum_{i=1}^{N}
  \mathbb{I}\!\left[y_{i}^{\mathrm{task}}=1\;\land\;
  y_{i}^{\mathrm{contact}}=1\right].
  \label{eq:csr}
\end{equation}
Here, $y_{i}^{\mathrm{task}}$ and $y_{i}^{\mathrm{contact}}$ are binary indicators
of task completion and contact validity for trial $i$, respectively;
$\mathbb{I}[\cdot]$ is the indicator function, and $N$ is the number of
independent trials. Contact validity is determined automatically from synchronized
tactile measurements using task-specific thresholds fixed before evaluation. A
trial is contact-invalid if required contact is lost or the contact force exceeds
the prescribed threshold. We run 20
real-robot trials per method per task, using the same initial-state
randomization, reset procedure, and stopping rule across methods. Four-task averages weight tasks
equally (80 trials per method) and are computed before rounding.
\subsubsection{Implementation Details}

For each backbone in Table~\ref{tab:main_results} (DP and $\pi_{0.5}$),
evaluations with and without ForeTac share the same demonstrations, base-policy
observation streams, action representation, frozen base-policy checkpoint, and
evaluation protocol.
TacVAE and Foresight are pretrained on a unified corpus of 5,969
timestamp-aligned episodes containing approximately 2.1 million marker frames.
TacVAE uses eight-frame marker histories with stride two, whereas Foresight uses
temporally aligned future action and tactile sequences. Episode identifiers are
split before temporal windows are constructed, and tactile normalization
statistics are computed from the training split and reused downstream.

We use
AdamW with (learning rate, batch size, epochs) set to
($10^{-4}$, 512, 150) for TacVAE, ($4\times10^{-5}$, 16, 100) for Foresight,
($10^{-4}$, 64, up to 1,000) for Diffusion Policy, and ($3\times10^{-4}$,
128, 40) for each scorer.
Foresight and the scorers use weight decay $10^{-4}$ and gradient clipping at
1.0; Diffusion Policy uses an exponential moving average. The policy uses global
and wrist RGB observations, the current joint state, a two-step observation
horizon, a 16-step action horizon, and 100 diffusion training steps.
Each scorer uses 100 episodes labeled good and 120 labeled bad, with an 80/20
episode-level split and outcome-balanced sampling. At deployment, the diffusion backbone uses 30 sampling
steps, applies guidance at ten steps with scale 0.003, executes the first eight
actions of the generated chunk, and then replans from a new observation.

\subsection{Main Results}

We first evaluate whether predictive contact guidance improves task completion
and physically acceptable contact execution over vision-only, direct tactile
fusion, and reactive tactile baselines. Table~\ref{tab:main_results} reports SR
and CSR over the four real-robot tasks under the protocol in Sec.~IV-A, with
separate blocks for the primary comparison and flow-matching evaluation.
For the DP and $\pi_{0.5}$ backbones, we compare task performance with and
without ForeTac.

\begin{table*}[t]
\caption{Real-robot task success rate (SR) and contact success rate (CSR), in
percent. Average is the mean over four tasks; best results within each block are
bold.}
\label{tab:main_results}
\centering
\setlength{\tabcolsep}{2.2pt}
\begin{tabular*}{\textwidth}{@{\extracolsep{\fill}}llcccccccccc@{}}
\toprule
\multirow{2}{*}{Tactile use} & \multirow{2}{*}{Method} &
\multicolumn{2}{c}{Board} & \multicolumn{2}{c}{Vase} &
  \multicolumn{2}{c}{Card} & \multicolumn{2}{c}{Chip} &
  \multicolumn{2}{c}{Average} \\
\cmidrule(lr){3-4}\cmidrule(lr){5-6}\cmidrule(lr){7-8}
  \cmidrule(lr){9-10}\cmidrule(lr){11-12}
  & & SR & CSR & SR & CSR & SR & CSR & SR & CSR & SR & CSR \\
\midrule
\multicolumn{12}{l}{\textit{Primary comparison}} \\
\multirow{2}{*}{No touch}
  & DP & 50 & 25 & 55 & 35 & 10 & 0 & 30 & 20 & 36.3 & 20.0 \\
  & RDT & 65 & 55 & 65 & 55 & 20 & 5 & 50 & 45 & 50.0 & 40.0 \\
\midrule
\multirow{2}{*}{Current touch}
  & DP + Tactile Concat & 60 & 50 & 60 & 50 & 35 & 30 & 40 & 35 & 48.8 & 41.3 \\
  & RDP & 70 & 60 & 65 & 55 & 45 & 35 & 55 & 45 & 58.8 & 48.8 \\
\midrule
Predictive touch & \textbf{DP + ForeTac} & \textbf{85} & \textbf{80} &
\textbf{70} & \textbf{65} & \textbf{55} & \textbf{50} &
  \textbf{75} & \textbf{70} & \textbf{71.3} & \textbf{66.3} \\
\midrule
\multicolumn{12}{l}{\textit{Flow-matching evaluation}} \\
No touch & $\pi_{0.5}$ & 55 & 45 & 60 & 45 & 25 & 20 & 45 & 35 & 46.3 & 36.3 \\
Predictive touch & $\boldsymbol{\pi}_{0.5}$ \textbf{+ ForeTac} &
  \textbf{80} & \textbf{75} & \textbf{70} & \textbf{65} &
  \textbf{50} & \textbf{45} & \textbf{70} & \textbf{65} &
  \textbf{67.5} & \textbf{62.5} \\
\bottomrule
\end{tabular*}
\end{table*}

DP + ForeTac is best on all four primary tasks, averaging 71.3\% SR and 66.3\%
CSR. It improves over DP by 35.0 and 46.3 percentage points and over RDP by
12.5 and 17.5 percentage points, while also exceeding direct current-tactile
concatenation on every task. These results show gains in both task success and
contact success across the four tasks.

\subsection{Ablation Studies}

We examine the contributions of tactile representation, predicted-contact quality,
and generation-time guidance by comparing three variants with full \textbf{ForeTac}.
\textbf{ForeTac w/ ResNet}
replaces the current-tactile encoder with a frozen ResNet-18~\cite{he2016resnet}
and a learned linear adapter while retaining TacVAE latents as future targets;
\textbf{Prediction Auxiliary Loss} adds tactile prediction to DP but omits
contact-quality scoring and deployment-time guidance; and \textbf{ForeTac w/
Predictive Reranking} samples a fixed number of candidates and executes the
highest-scoring one without propagating the quality gradient into sampling. All
variants use the same action space, prediction horizon, and trial protocol;
reranking additionally uses the same frozen Foresight, scorer,
candidate-generation protocol, and base-policy configuration.

\begin{table}[t]
\caption{Ablation of tactile representation and the use of predicted contact
quality. Task-specific cells report SR/CSR (\%); Average is the mean over four
tasks. Best results are bold.}
\label{tab:ablation}
\centering
\setlength{\tabcolsep}{1.8pt}
\begin{tabular*}{\columnwidth}{@{\extracolsep{\fill}}lccccc@{}}
\toprule
Variant & Board & Vase & Card & Chip & Average \\
\midrule
ForeTac w/ ResNet & 55/55 & 55/40 & 35/10 & 55/50 & 50.0/38.8 \\
Prediction Auxiliary Loss & 60/55 & 65/55 & 40/35 & 55/50 & 55.0/48.8 \\
ForeTac w/ Predictive Reranking & 65/60 & 65/60 & 45/35 & 60/55 & 58.8/52.5 \\
\textbf{ForeTac} & \textbf{85/80} & \textbf{70/65} & \textbf{55/50} &
\textbf{75/70} & \textbf{71.3/66.3} \\
\bottomrule
\end{tabular*}
\end{table}

As shown in Table~\ref{tab:ablation}, the complete method achieves the highest
SR and CSR on every task. Using the ResNet input encoder
reduces average SR and CSR by 21.3 and 27.5 percentage points, respectively,
supporting the role of the learned contact representation within the complete
pipeline.
Predictive reranking matches or improves the auxiliary-loss variant
but remains below full ForeTac on all tasks, with average SR and CSR gaps of
12.5 and 13.8 percentage points, respectively. This gap highlights the benefit
of guiding action generation with predicted-contact gradients, rather than
merely reranking generated actions.
As a control, we replaced the future-tactile rollout with a direct scorer of
current tactile history, proprioception, and candidate actions, trained on the
same good/bad labels and evaluated under the same diffusion guidance schedule.
This direct current-contact guidance achieved 60.0\% SR and 51.3\% CSR over the
four tasks, compared with 71.3\% and 66.3\% for ForeTac, indicating the benefit
of evaluating predicted future contact.

\subsection{Component Analysis}

\subsubsection{Tactile Representation}

\paragraph*{\normalfont\bfseries Reconstruction Fidelity}

We first examine whether TacVAE compactly represents the marker-displacement
sequences required by the downstream predictor and contact-quality model. We
compare it with PCA, PointNet-AE, and Conv-AE on the same held-out
task-specific marker-displacement windows from the four tasks in
Table~\ref{tab:tactile_representation}, using reconstruction MAE and cosine
similarity. Metrics use training-split channel statistics and are averaged
first over markers and frames and then over windows.

\begin{table}[t]
\caption{Tactile reconstruction across representation methods. Lower MAE and
higher cosine similarity are better; best results are bold.}
\label{tab:tactile_representation}
\centering
\setlength{\tabcolsep}{1.35pt}
\begin{tabular*}{\columnwidth}{@{\extracolsep{\fill}}lcccccccc@{}}
\toprule
\multirow{2}{*}{Encoder} & \multicolumn{2}{c}{Board} &
\multicolumn{2}{c}{Vase} & \multicolumn{2}{c}{Card} &
\multicolumn{2}{c}{Chip} \\
\cmidrule(lr){2-3}\cmidrule(lr){4-5}\cmidrule(lr){6-7}\cmidrule(lr){8-9}
& MAE & Cos. & MAE & Cos. & MAE & Cos. & MAE & Cos. \\
\midrule
PCA & 0.040 & 0.995 & 0.035 & \textbf{0.995} & 0.036 & 0.980 & 0.026 & 0.965 \\
PointNet-AE & 0.048 & 0.956 & \textbf{0.027} & 0.955 & 0.044 & 0.992 & 0.032 & \textbf{0.990} \\
Conv-AE & 0.055 & 0.997 & 0.042 & 0.994 & 0.104 & \textbf{0.999} & 0.036 & 0.959 \\
TacVAE & \textbf{0.039} & \textbf{0.998} & 0.037 & 0.975 &
\textbf{0.031} & \textbf{0.999} & \textbf{0.015} & 0.899 \\
\bottomrule
\end{tabular*}
\end{table}

Table~\ref{tab:tactile_representation} shows that TacVAE has the lowest MAE on
Board, Card, and Chip. Its cosine similarity is highest on Board and Card but
lower on Vase and Chip; Table~\ref{tab:ablation} tests its contribution to
closed-loop performance.

\subsubsection{Action-Conditioned Tactile Foresight}

\paragraph{\normalfont\bfseries Action-Conditioning Validation}

We test whether Foresight uses its candidate-action input rather than
extrapolating tactile history alone. On 20,055 matched windows from 23 held-out
Board episodes, with fixed observations and recorded tactile targets, the same
predictor is evaluated with an aligned candidate action or an action-masked
control. Each forward pass predicts the complete 16-step tactile sequence, and
errors are averaged over all future steps and latent entries
(Table~\ref{tab:action_conditioning}).

\begin{table}[t]
\caption{Action-conditioning validation on matched held-out observation
windows. Lower prediction errors are better.}
\label{tab:action_conditioning}
\centering
\setlength{\tabcolsep}{2.8pt}
\begin{tabular*}{\columnwidth}{@{\extracolsep{\fill}}lcc@{}}
\toprule
Configuration & Latent MAE & Marker error \\
\midrule
Aligned candidate action & 0.0589 & 0.2832~px \\
Action-masked control & 0.0941 & 0.5209~px \\
\bottomrule
\end{tabular*}
\end{table}

The action-masked control increases latent MAE from 0.0589 to 0.0941 and marker
error from 0.2832 to
0.5209 px. This degradation indicates that the predictor uses candidate-action
information.

\paragraph{\normalfont\bfseries Effective Foresight Horizon and Control Utility}

We test how the effective foresight prefix affects prediction accuracy and
closed-loop control utility by evaluating the fixed 16-step Foresight predictor
with valid prefixes $L\in\{1,4,8,12,16\}$. For $L<16$, the remaining predicted states are masked
before scoring and guidance, while the scorer remains fixed across settings.
Prediction errors are averaged over the first $L$ steps on held-out evaluation
windows.
SR/CSR are measured in the corresponding robot rollouts; all other settings
remain unchanged.

\begin{table}[t]
\caption{Prediction quality and Board rollout SR/CSR across effective foresight
horizons $L$. The predictor outputs 16 steps; only the first $L$ steps are
valid for scoring and guidance, and the remaining positions are masked.}
\label{tab:foresight_horizon}
\centering
\setlength{\tabcolsep}{3.0pt}
\begin{tabular}{cccccc}
\toprule
$L$ & Latent MAE & Latent cos. & Marker error & SR & CSR \\
\midrule
1  & 0.0485 & 0.9989 & 0.1973 px & 60\% & 50\% \\
4  & 0.0498 & 0.9987 & 0.2147 px & 65\% & 50\% \\
8  & 0.0534 & 0.9984 & 0.2443 px & 75\% & 65\% \\
12 & 0.0563 & 0.9981 & 0.2650 px & 80\% & 65\% \\
16 & 0.0589 & 0.9978 & 0.2832 px & \textbf{85\%} & \textbf{80\%} \\
\bottomrule
\end{tabular}
\end{table}

Table~\ref{tab:foresight_horizon} reports the prefix-averaged prediction errors
and rollout SR/CSR under this fixed-scorer masking protocol. Prediction error
increases with the evaluated prefix length, whereas the reported rollout
performance improves: SR rises from 60\% to 85\% and CSR from 50\% to 80\%.
Under this protocol, $L=16$ performs best among the tested settings.

\subsubsection{Contact-Quality Modeling}

\begin{figure*}[!t]
  \centering
  \includegraphics[width=0.84\textwidth]{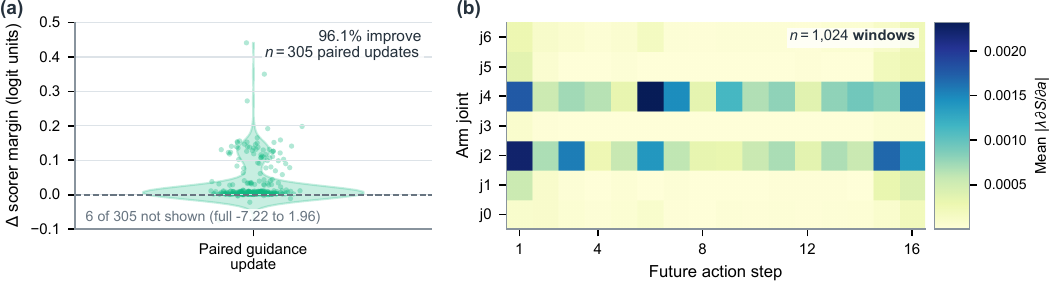}
  \caption{Offline diagnostics of predictive contact guidance. (a) Change
  in the predicted-contact quality margin for 305 paired guidance updates; 293
  (96.1\%) are positive. (b) Mean $|\lambda\,\partial S/\partial a|$
  ($\lambda=0.003$) across seven arm joints and future action steps over 1,024
  held-out windows.}
  \label{fig:guidance_diagnostic}
\end{figure*}

\paragraph*{\normalfont\bfseries Contact-Quality Ranking on Recorded and Predicted Future Touch}

Effective use of the contact-quality objective requires two properties: it must
rank good contact above bad contact, and predicted tactile
trajectories must preserve this ordering for guidance.

We first test whether the task-specific scorer ranks good contact above bad
contact in held-out offline rollouts. For each task, tactile windows are
extracted from the task-relevant contact phase of each rollout. Each pair
contains one window labeled good and one labeled bad; the preferred order is
defined by the corresponding task-specific physical outcome labels. The
recorded marker windows are encoded by the same frozen TacVAE used
to train the scorer. For a set of pairs $\mathcal{P}$, preference-order
accuracy~\cite{tian2026badbehavior} is
\begin{equation}
  A_{\mathrm{pref}}=\frac{1}{|\mathcal{P}|}
  \sum_{(i,j)\in\mathcal{P}}\mathbb{I}[s_i^{+}>s_j^{-}],
  \label{eq:preference_accuracy}
\end{equation}
where $s_i^{+}$ and $s_j^{-}$ are the quality margins $S_\psi$ evaluated on the
good and bad future contact trajectories, respectively.

\begin{table}[t]
\caption{Recorded and predicted future-touch ranking on held-out contact-phase
pairs. Recorded reports correct pairs and preference-order accuracy; predicted
replaces recorded latents with aligned Foresight predictions. $\rho$ is the rank
correlation between score sets.}
\label{tab:contact_quality_ranking}
\centering
\setlength{\tabcolsep}{2.3pt}
\begin{tabular*}{\columnwidth}{@{\extracolsep{\fill}}lcccc@{}}
\toprule
Task & Pairs & Recorded & Predicted & $\rho$ \\
\midrule
Board  & 1980 & 1856/1980 (93.7\%) & 95.1\% & 0.9368 \\
Vase   & 1012 & 936/1012 (92.5\%) & 91.1\% & 0.9728 \\
Card   & 1408 & 1292/1408 (91.8\%) & 94.2\% & 0.9321 \\
Chip   & 1240 & 1062/1240 (85.6\%) & 89.5\% & 0.9326 \\
\midrule
Pooled & 5640 & 5146/5640 (91.2\%) & 92.9\% & -- \\
\bottomrule
\end{tabular*}
\end{table}

Using the same frozen scorer and pairs, we replace each
recorded latent trajectory with its time-aligned Foresight prediction.
Table~\ref{tab:contact_quality_ranking} reports 85.6--93.7\% recorded-input
accuracy across tasks and 91.2\% over 5,640 pairs; predicted inputs achieve
89.5--95.1\% and 92.9\% pooled accuracy, with task-wise rank correlations of
$\rho=0.9321$--$0.9728$. The close accuracies and high rank correlations indicate
that predicted latent trajectories preserve the contact-quality ordering on
these held-out contact windows.

Figure~\ref{fig:guidance_diagnostic} diagnoses the generation-time guidance
signal: 293 of 305 paired updates (96.1\%) increase the predicted-contact
quality margin, and the gradient magnitudes are measured across the seven joints
and future action steps. These offline diagnostics support the quality-gradient
signal used during sampling.

\subsection{Cross-Architecture Analysis}

We test portability by attaching the same ForeTac representation, foresight
predictor, contact-quality objective, and guidance procedure to the frozen
flow-matching $\pi_{0.5}$ backbone; only the action-generation update changes.
Using the same initial-state randomization, 20-trial protocol, stopping rules,
and success criteria, with guidance applied during the final two of ten
integration steps, the flow-matching block of Table~\ref{tab:main_results}
shows gains of 21.3 and 26.3 percentage points in average SR and CSR
(46.3$\rightarrow$67.5\% and 36.3$\rightarrow$62.5\%). Together with the DP
block, these results show gains with both evaluated action-generation
backbones.

\subsection{Computational Cost}

We quantify the replanning overhead introduced by predictive contact guidance.
On an NVIDIA RTX 4090, batch-one replanning takes $261.26\pm5.16$~ms for
ForeTac and $132.46\pm1.39$~ms for DP.
These measurements use 50 warm-up replans followed by 200 synchronized timed
replans. ForeTac therefore adds 128.80~ms per replan, mainly due to repeated
tactile prediction, scoring, and gradient computation during guided sampling.

\section{Conclusion and Limitations}

ForeTac turns action-conditioned future touch into a differentiable signal for
generative action formation. Across four real-robot tasks, it improves average
SR/CSR over Diffusion Policy from 36.3\%/20.0\% to
71.3\%/66.3\%, and also improves both metrics with the flow-matching
$\pi_{0.5}$ backbone. The ablations and ranking analyses support the roles of
the learned tactile representation, action-conditioned prediction, and
generation-time gradient guidance, while predicted latent trajectories retain
contact-quality ordering on held-out contact windows.

The current system uses task-specific contact-quality models trained from good
and bad physical interactions, and repeated gradient computation increases
replanning latency. Future work will extend contact-quality modeling across
tasks and improve the efficiency and uncertainty handling of gradient-guided
replanning.

\bibliographystyle{IEEEtran}
\bibliography{references}

\end{CJK*}
\end{document}